\documentclass[11pt,a4paper]{article}
\usepackage[hyperref]{acl2021}
\pdfoutput=1 
\usepackage{times}
\usepackage{latexsym}
\usepackage[T1]{fontenc}
\usepackage[utf8]{inputenc}
\usepackage{microtype}
\usepackage{amsmath}
\usepackage{amssymb}
\usepackage{fontawesome}
\usepackage{graphics}
\usepackage{graphicx}
\usepackage{subfigure}
\usepackage{booktabs}
\usepackage{upgreek}
\usepackage{fontawesome}
\usepackage{siunitx}
\usepackage{microtype}

\aclfinalcopy 
\def\aclpaperid{3007} 

\title{MedNLI Is Not Immune:\\ Natural Language Inference Artifacts in the Clinical Domain}

\author{Christine Herlihy \\
  Department of Computer Science \\
  University of Maryland \\
  College Park, MD \\
  \texttt{cherlihy@cs.umd.edu} \\\And
  Rachel Rudinger \\
  Department of Computer Science \\
  University of Maryland \\
  College Park, MD \\
  \texttt{rudinger@umd.edu} \\}

\date{}

\begin{document}
\maketitle
\begin{abstract}
Crowdworker-constructed natural language inference (NLI) datasets have been found to contain statistical artifacts associated with the annotation process that allow hypothesis-only classifiers to achieve better-than-random performance~\citep{poliak-etal-2018-hypothesis, gururangan2018annotation,tsuchiya-2018-performance}. We investigate whether MedNLI, a physician-annotated dataset with premises extracted from clinical notes, contains such artifacts~\citep{romanov2018lessons}.

We find that entailed hypotheses contain generic versions of specific concepts in the premise, as well as modifiers related to responsiveness, duration, and probability. Neutral hypotheses feature conditions and behaviors that co-occur with, or cause, the condition(s) in the premise. Contradiction hypotheses feature explicit negation of the premise and implicit negation via assertion of good health. Adversarial filtering demonstrates that performance degrades when evaluated on the \emph{difficult} subset. We provide partition information and recommendations for alternative dataset construction strategies for knowledge-intensive domains.
\end{abstract}

\section{Introduction}
In the clinical domain, the ability to conduct natural language inference (NLI) on unstructured, domain-specific texts such as patient notes, pathology reports, and scientific papers, plays a critical role in the development of predictive models and clinical decision support (CDS) systems.  

Considerable progress in domain-agnostic NLI has been facilitated by the development of large-scale, crowdworker-constructed datasets, including the Stanford Natural Language Inference corpus (SNLI), and the Multi-Genre Natural Language Inference (MultiNLI) corpus~\citep{bowman2015large, williams2017broadcoverage}. MedNLI is a similarly-motivated, healthcare-specific dataset created by a small team of  physician-annotators in lieu of crowdworkers, due to the extensive domain expertise required~\citep{romanov2018lessons}.

\citet{poliak-etal-2018-hypothesis}, \citet{gururangan2018annotation}, \citet{tsuchiya-2018-performance}, and \citet{mccoy2019right} empirically demonstrate that SNLI and MultiNLI contain lexical and syntactic annotation artifacts that are disproportionately associated with specific classes, allowing a hypothesis-only classifier to significantly outperform a majority-class baseline model. The presence of such artifacts is hypothesized to be partially attributable to the priming effect of the example hypotheses provided to crowdworkers at annotation-time. \citet{romanov2018lessons} note that a hypothesis-only baseline is able to outperform a majority class baseline in MedNLI, but they do not identify specific artifacts.

We confirm the presence of annotation artifacts in MedNLI and proceed to identify their lexical and semantic characteristics. We then conduct adversarial filtering to partition MedNLI into \emph{easy} and \emph{difficult} subsets~\citep{sakaguchi2020winogrande}. We find that performance of off-the-shelf \texttt{fastText}-based hypothesis-only and hypothesis-plus-premise classifiers is lower on the \emph{difficult} subset than on the \emph{full} and \emph{easy} subsets~\citep{joulin2016bag}. We provide partition information for downstream use, and conclude by advocating alternative dataset construction strategies for knowledge-intensive domains.\footnote{See {https://github.com/crherlihy/clinical\_nli\_artifacts} for code and partition ids.}

\section{The MedNLI Dataset}
MedNLI is domain-specific evaluation dataset inspired by general-purpose NLI datasets, including SNLI and MultiNLI~\citep{romanov2018lessons, bowman2015large, williams2017broadcoverage}. Much like its predecessors, MedNLI consists of premise-hypothesis pairs, in which the premises are drawn from the \texttt{Past Medical History} sections of a randomly selected subset of de-identified clinical notes contained in MIMIC-III~\citep{johnson2016mimicIII, goldberger2000physiobank}. MIMIC-III  was created from the records of adult and neonatal intensive care unit (ICU) patients. As such, complex and clinically severe cases are disproportionately represented, relative to their frequency of occurrence in the general population. 

Physician-annotators were asked to write a \textit{definitely true}, \textit{maybe true}, and \textit{definitely false} set of hypotheses for each premise, corresponding to \textit{entailment}, \textit{neutral} and \textit{contradiction} labels, respectively. The resulting dataset has cardinality: $n_{\text{train}} = 11232; \ n_{\text{dev}} = 1395; \ n_{\text{test}} = 1422$.

\section{MedNLI Contains Artifacts}
\label{sec:mednliContainsArtifacts}
To determine whether MedNLI contains annotation artifacts that may artificially inflate the performance of models trained on this dataset, we train a simple, premise-unaware, \texttt{fastText} classifier to predict the label of each premise-hypothesis pair, and compare the performance of this classifier to a majority-class baseline, in which all training examples are mapped to the most commonly occurring class label~\citep{joulin2016bag, poliak-etal-2018-hypothesis, gururangan2018annotation}. Note that since annotators were asked to create an entailed, contradictory, and neutral hypothesis for each premise, MedNLI is class-balanced. Thus, in this setting, a majority class baseline is equivalent to choosing a label uniformly at random for each training example. 

The micro F1-score achieved by the \texttt{fastText} classifier significantly exceeds that of the majority class baseline, confirming the findings of \citet{romanov2018lessons}, who report a micro-F1 score of 61.9 but do not identify or analyze artifacts:
\begin{table}[!bth]
\centering
\small
\begin{tabular}{lcc}
\hline
{} &   \textbf{dev} &  \textbf{test} \\
\hline
majority class &  33.3 &  33.3 \\
\texttt{fastText} &  \textbf{64.8} &  \textbf{62.6} \\
\hline
\end{tabular}
\caption{Performance (micro F1-score) of the \texttt{fastText} hypothesis-only classifier.}
\label{tab:fastTextF1}
\end{table}

As the confusion matrix for the test set shown in \hyperref[tab:confusionMatrix]{Table 2} indicates, the \texttt{fastText} model is most likely to misclassify entailment as neutral, and neutral and contradiction as entailment. Per-class precision and recall on the test set are highest for contradiction (73.2; 72.8) and lowest for entailment (56.7; 53.8).

\begin{table}[!htb]
\resizebox{\columnwidth}{!}{%
\centering
\begin{tabular}{lccc}
\hline
{} &  entailment &  neutral & contradiction \\
\hline
entailment &  \textbf{255} &  151 &   68 \\
neutral &  126 &  \textbf{290} &   58 \\
contradiction &   69 &   60 &  \textbf{345} \\
\hline
\end{tabular}%
}
\caption{Confusion matrix for \texttt{fastText} classifier.}
\label{tab:confusionMatrix}
\end{table}

\section{Characteristics of Clinical Artifacts}
In this section, we conduct class-specific lexical analysis to identify the clinical and domain-agnostic characteristics of annotation artifacts associated with each set of hypotheses in MedNLI. 

\subsection{Preprocessing}
We cast each hypothesis string in the MedNLI training dataset to lowercase. We then use a \texttt{scispaCy} model pre-trained on the \texttt{en\_core\_sci\_lg} corpus for tokenization and clinical named entity recognition (CNER)~\citep{neumann2019scispacy}. One challenge associated with clinical text, and scientific text more generally, is that semantically meaningful entities often consist of spans rather than single tokens. To mitigate this issue during lexical analysis, we map each multi-token entity to a single-token representation, where sub-tokens are separated by underscores. 

\subsection{Lexical Artifacts}
\label{sec:lexicalArtifacts}
Following \citet{gururangan2018annotation}, to identify tokens that occur disproportionately in hypotheses associated with a specific class, we compute token-class pointwise mutual information (PMI) with add-50 smoothing applied to raw counts, and a filter to exclude tokens appearing less than five times in the overall training dataset. Table \hyperref[tab:wordChoice]{3} reports the top 15 tokens for each class.
\begin{equation}
\noindent
\nonumber
        \texttt{PMI}(\text{token, class}) = log_2 \frac{p(\text{token, class})}{p(\text{token}, \cdot ) p(\cdot, \text{class})}
\end{equation}

\begin{table*}[!htb]
\small
\centering
\begin{tabular}{lclclc}
\toprule
entailment & \% &               neutral & \% &        contradiction & \% \\
\midrule
just & 0.25\% &     cardiogenic\_shock & 0.33\% & no\_history\_of\_cancer & 0.27\% \\
high\_risk & 0.26\% &           pelvic\_pain & 0.30\% &         no\_treatment & 0.27\% \\
pressors & 0.25\% &            joint\_pain & 0.30\% &     normal\_breathing & 0.27\% \\
possible & 0.26\% &          brain\_injury & 0.32\% &  no\_history\_of\_falls & 0.27\% \\
elevated\_blood\_pressure & 0.26\% &              delerium & 0.30\% &  normal\_heart\_rhythm & 0.28\% \\
responsive & 0.25\% & intracranial\_pressure & 0.30\% &               health & 0.26\% \\
comorbidities & 0.26\% &               smoking & 0.42\% &       normal\_head\_ct & 0.26\% \\
spectrum & 0.27\% &               obesity & 0.41\% &        normal\_vision & 0.26\% \\
steroid\_medication & 0.25\% &                   tia & 0.32\% &  normal\_aortic\_valve & 0.27\% \\
longer & 0.26\% &              acquired & 0.31\% &          bradycardic & 0.26\% \\
history\_of\_cancer & 0.26\% &           head\_injury & 0.31\% &  normal\_blood\_sugars & 0.27\% \\
broad & 0.26\% &                 twins & 0.30\% &    normal\_creatinine & 0.28\% \\
frequent & 0.25\% &             fertility & 0.30\% &       cancer\_history & 0.26\% \\
failed & 0.26\% &                statin & 0.30\% &              cardiac & 0.33\% \\
medical & 0.29\% &          acute\_stroke & 0.30\% &         normal\_chest & 0.28\% \\
\bottomrule
\end{tabular}
\caption{Top 15 tokens by \texttt{PMI}(token, class); \% of \emph{class} training examples that contain the token.}
\label{tab:wordChoice}
\end{table*}

\paragraph{Entailment} 
Entailment hypotheses are characterized by tokens about: (1) patient status and response to treatment (e.g., \emph{responsive}; \emph{failed}; \emph{longer} as in \emph{no longer intubated}); (2) medications and procedures which are common among ICU patients (e.g., \emph{broad\_spectrum}; \emph{antibiotics}; \emph{pressors};  \emph{steroid\_medication}; \emph{underwent}; \emph{removal}); (3) generalized versions of specific words in the premise (e.g., \emph{comorbidities}; \emph{multiple\_medical\_problems}), which \citet{gururangan2018annotation} also observe in SNLI; and (4) modifiers related to duration, frequency, or probability (e.g., \emph{frequent}, \emph{possible}, \emph{high\_risk}). 
\paragraph{Neutral}
Neutral hypotheses feature tokens related to: (1) chronic and acute clinical conditions (e.g., \emph{obesity}; \emph{joint\_pain}; \emph{brain\_injury}); (2) clinically relevant behaviors (e.g., \emph{smoking}; \emph{alcoholic}; \emph{drug\_overdose}); and (3) gender and reproductive status (e.g., \emph{fertility}; \emph{pre\_menopausal}).
Notably, the most discriminative conditions tend to be commonly occurring within the general population and generically stated, rather than rare and specific. This presumably contributes to the relative difficulty that the hypothesis-only \texttt{fastText} model has distinguishing between the entailment and neutral classes. 

\paragraph{Contradiction}
Contradiction hypotheses are characterized by tokens that convey normalcy and good health. Lexically, such sentiment manifests as: (1) explicit negation of clinical severity, medical history, or in-patient status (e.g., \emph{denies\_pain}; \emph{no\_treatment}; \emph{discharged\_home}), or (2) affirmation of clinically unremarkable findings (e.g., \emph{normal\_heart\_rhythm}; \emph{normal\_blood\_sugars}), which would generally be rare among ICU patients. This suggests a heuristic of inserting negation token(s) to contradict the premise, which \citet{gururangan2018annotation} also observe in SNLI.

\subsection{Syntactic Artifacts}
\paragraph{Hypothesis Length}
In contrast to \citet{gururangan2018annotation}'s finding
that entailed hypotheses in SNLI tend to be shorter while neutral hypotheses tend to be longer, hypothesis sentence length does not appear to play a discriminatory role in MedNLI, regardless of whether we consider merged- or separated-token representations of multi-word entities, as illustrated by Table \hyperref[fig:hypLenTable]{4}:

\begin{table}[!htb]
\label{fig:hypLenTable}
\resizebox{\columnwidth}{!}{%
\begin{tabular}{l|c|c|c|c|c|c|}
\cline{2-7}
                                                      & \multicolumn{2}{c|}{\textbf{entailment}} & \multicolumn{2}{c|}{\textbf{neutral}} & \multicolumn{2}{c|}{\textbf{contradiction}} \\ \cline{2-7} 
                                                      & mean               & median              & mean             & median             & mean                & median                \\ \hline
\multicolumn{1}{|c|}{\textbf{separate}} & 5.6                & 5.0                 & 5.2              & 5.0                & 5.6                 & 5.0                   \\ \hline
\multicolumn{1}{|c|}{\textbf{merged}}   & 5.3                & 5.0                 & 4.9              & 5.0                & 5.3                 & 5.0                   \\ \hline
\end{tabular}%
}
\caption{Average and median hypothesis length by class and entity representation.}
\end{table}

\section{Physician-Annotator Heuristics} 
\label{sec:annotatorHeuristics}
In this section, we re-introduce premises to our analysis to evaluate a set of hypotheses regarding latent, class-specific annotator heuristics. If annotators \emph{do} employ class-specific heuristics, we should expect the semantic contents, $\upvarphi$, of a given hypothesis, $h \in \mathcal{H}$, to be influenced not only by the semantic contents of its associated premise, $p \in \mathcal{P}$, but also by the target class, $c \in \mathcal{C}$. 

To investigate, we identify a set of heuristics parameterized by $\upvarphi(p)$ and $c$, and characterized by the presence of a set of heuristic-specific Medical Subject Headings (MeSH) linked entities in the premise and hypothesis of each heuristic-satisfying example. These heuristics are described below; specific MeSH features are detailed in the \hyperref[sec:appendix]{Appendix}. 

\paragraph{Hypernym Heuristic} This heuristic applies when the premise contains clinical condition(s), medication(s), finding(s), procedure(s) or event(s), the target class is \emph{entailment}, and the generated hypothesis contains term(s) that can be interpreted as super-types for a subset of elements in the premise (e.g., clindamycin \texttt{<:} antibiotic). 

\paragraph{Probable Cause Heuristic} This heuristic applies when the premise contains clinical condition(s), the target class is \emph{neutral}, and the generated hypothesis provides a plausible, often subjective or behavioral, causal explanation for the condition, finding, or event described in the premise (e.g., associating altered mental status with drug overdose).

\paragraph{Everything Is Fine Heuristic} This heuristic applies when the premise contains condition(s) or finding(s), the target class is \emph{contradiction}, and the generated hypothesis negates the premise or asserts unremarkable finding(s). This can take two forms: repetition of premise content plus negation, or inclusion of modifiers that convey good health.

\paragraph{Analysis}
We conduct a $\chi^2$ test for each heuristic to determine whether we are able to reject the null hypothesis that pattern-satisfying premise-hypothesis pairs are uniformly distributed over classes. 

\begin{table}[!htb]
\footnotesize
\centering
\begin{tabular}{lcll}
\toprule
         \textbf{heuristic} & $\chi^2$ &      \textbf{p-value} & \textbf{top class} \\
\midrule
          hypernym &           59.15 &     \num{1.4e-13}$\ddagger$ &     entail (45.2\%) \\
    probable cause &          111.05 &  \num{7.7e-25}$\ddagger$ &        neutral (57.8\%) \\
 everything fine &         874.71 &  \num{1.1e-190}$\ddagger$ &  contradict (83.8\%) \\
\bottomrule
\end{tabular}%
\caption{Results of $\chi^2$ test statistic by heuristic, computed using the combined MedNLI dataset \footnotesize{($\ddagger$ $p<0.001$, $\dagger$ $p<0.01$, * $p<0.5$).} Top class presented with \% of heuristic-satisfying pairs.}
\label{tab:chisquare}
\end{table}

The results support our hypotheses regarding each of the three heuristics. Notably, the percentage of heuristic-satisfying pairs accounted for by the top class is lowest for the \textsc{hypernym} hypothesis, which we attribute to the high degree of semantic overlap between entailed and neutral hypotheses. 

\section{Adversarial Filtering}
To mitigate the effect of clinical annotation artifacts, we employ \texttt{AFLite}, an adversarial filtering algorithm introduced by \citet{sakaguchi2020winogrande} and analyzed by \citet{bras2020adversarial}, to create \emph{easy} and \emph{difficult} partitions of MedNLI. 

\texttt{AFLite} requires distributed representations of the full dataset as input, and proceeds in an iterative fashion. At each iteration, an ensemble of $n$ linear classifiers are trained and evaluated on different random subsets of the data. A score is then computed for each premise-hypothesis instance, reflecting the number of times the instance is correctly labeled by a classifier, divided by the number of times the instance appears in any classifier's evaluation set. The top-$k$ instances with scores above a threshold, $\tau$, are filtered out and added to the \emph{easy} partition; the remaining instances are retained. This process continues until the size of the filtered subset is $< k$, or the number of retained instances is $< m$; retained instances constitute the \emph{difficult} partition.  

To represent the full dataset, we use $\texttt{fastText}_{\text{MIMIC-III}}$ embeddings, which have been pretrained on deidentified patient notes from MIMIC-III~\citep{romanov2018lessons, johnson2016mimicIII}. We represent each example as the average of its component token vectors.  

We proportionally adjust a subset of the hyperparameters used by \citet{sakaguchi2020winogrande} to account for the fact that MedNLI contains far fewer examples than \textsc{Winogrande}\footnote{MedNLI's training dataset contains $14049$ examples when the training, dev, and test sets are combined, while \textsc{Winogrande} contains $47$K after excluding the $6$K used for fine-tuning.}: specifically, we set the training size for each ensemble, $m$, to $5620$, which represents $\approx \frac{2}{5}$ of the MedNLI combined dataset. The remaining hyperparameters are unchanged: the ensemble consists of $n = 64$ logistic regression models, the filtering cutoff, $k$ = 500, and the filtering threshold $\tau =0.75$. 

We apply \texttt{AFLite} to two different versions of MedNLI: (1) $\mathcal{X}_{h,m}$: hypothesis-only, multi-token entities merged, and (2) $\mathcal{X}_{ph,m}$: premise and hypothesis concatenated, multi-token entities merged. \texttt{AFLIte} maps each version to an \emph{easy} and \emph{difficult} partition, which can in turn be split into training, dev, and test subsets. We report results for the \texttt{fastText} classifier trained on the original, hypothesis-only (hypothesis + premise) MedNLI training set, and evaluated on the \emph{full}, \emph{easy} and \emph{difficult} dev and test subsets of $\mathcal{X}_{h,m}$ ($\mathcal{X}_{ph,m}$), and observe that performance decreases on the \emph{difficult} partition:

\begin{table}[!htb]
\resizebox{\columnwidth}{!}{%
    \centering
    \begin{tabular}{lccccc}
    \toprule
    {} &           \textbf{model} & \textbf{eval dataset} & \textbf{full} &  \textbf{easy} ($\Delta$) &  \textbf{difficult} ($\Delta$) \\
    \midrule
    \textbf{no premise} &  majority class &            dev &  0.33 &  0.34 (+0.01) &  0.35 (+0.02) \\
    \textbf{no premise} &  majority class &        test &  0.33 &  0.35 (+0.02) &  0.37 (+0.04) \\
    \hline
    \textbf{no premise}  &        \texttt{fastText} &        dev &  0.65 &  0.67 (+0.02) &  0.46 (-0.19) \\
    \textbf{no premise}  &        \texttt{fastText} &         test &  0.63 &  0.65 (+0.02) &   0.4 (-0.23) \\
    \hline
    \textbf{with premise} &  majority class &          dev &  0.33 &  0.45 (+0.12) &  0.36 (+0.03) \\
    \textbf{with premise} &  majority class &         test &  0.33 &  0.48 (+0.15) &  0.37 (+0.04) \\
    \hline
    \textbf{with premise} &        \texttt{fastText} &          dev &  0.53 &   0.6 (+0.07) &   0.43 (-0.1) \\
    \textbf{with premise} &        \texttt{fastText} &         test &  0.51 &  0.55 (+0.04) &   0.4 (-0.11) \\
    \bottomrule
    \end{tabular}
    }
    \caption{Performance (micro F1-score) for the majority class baseline and \texttt{fastText} classifiers, with and without premise, by partition (e.g., \emph{full, easy, difficult}).}
    \label{tab:afliteRes}
\end{table}

\section{Discussion}
\subsection{MedNLI is Not Immune from Artifacts} 
In this paper, we demonstrate that MedNLI suffers from the same challenge associated with annotation artifacts that its domain-agnostic predecessors have encountered: namely, NLI models trained on $\{$Med, S, Multi$\}$NLI can perform well even without access to the training examples' premises, indicating that they often exploit shallow heuristics, with negative implications for out-of-sample generalization. 

Interestingly, many of the high-level lexical characteristics identified in MedNLI can be considered domain-specific variants of the more generic, class-specific patterns identified in SNLI. This observation suggests that a set of abstract design patterns for inference example generation exists across domains, and may be reinforced by the prompts provided to annotators. Creative or randomized priming, such as \citet{sakaguchi2020winogrande}~'s use of anchor words from WikiHow articles, may help to decrease reliance on such design patterns, but it appears unlikely that they can be systematically sidestepped without introducing new, ``corrective'' artifacts. 

\subsection{A Prescription for Dataset Construction}
To mitigate the risk of performance overestimation associated with annotation artifacts, \citet{zellers2019hellaswag} advocate adversarial dataset construction, such that benchmarks will co-evolve with language models. This may be difficult to scale in knowledge-intensive domains, as expert validation of adversarially generated benchmarks is typically required. Additionally, in high-stakes domains such as medicine, information-rich inferences should be preferred over correct but trivial inferences that time-constrained expert annotators may be rationally incentivized to produce, because entropy-reducing inferences are more useful for downstream tasks. 

We advocate the adoption of a mechanism design perspective, so as to develop modified annotation tasks that reduce the cognitive load placed on expert annotators while incentivizing the production of domain-specific NLI datasets with high downstream utility~\citep{ho2015incentivizing, liu2017machine}. An additional option is to narrow the generative scope by defining a set of inferences deemed to be useful for a specific task. Annotators can then map (premise, relation) tuples to relation-satisfying, potentially fuzzy subsets of this pool of useful inferences, or return partial functions when more information is needed. 

\section{Ethical Considerations}
When working with clinical data, two key ethical objectives include: (1) the preservation of patient privacy, and (2) the development of language and predictive models that benefit patients and providers to the extent possible, without causing undue harm. With respect to the former, MedNLI's premises are sampled from de-identified clinical notes contained in MIMIC-III~\cite{goldberger2000physiobank, johnson2016mimicIII}, and the hypotheses generated by annotators do not refer to specific patients, providers, or locations by name. MedNLI requires users to complete Health Insurance Portability and Accountability Act (HIPAA) training and sign a data use agreement prior to being granted access, which we have complied with. 

Per MedNLI's data use agreement requirements, we do not attempt to identify any patient, provider, or institution mentioned in the de-identified corpus. Additionally, while we provide AFLite \emph{easy} and \emph{difficult} partition information for community use in the form of split-example ids and a checksum, we do not share the premise or hypothesis text associated with any example. Interested readers are encouraged to complete the necessary training and obtain credentials so that they can access the complete dataset~\cite{romanov2018lessons, goldberger2000physiobank}.

With respect to benefiting patients, the discussion of natural language artifacts we have presented is intended to encourage clinical researchers who rely on (or construct) expert-annotated clinical corpora to train domain-specific language models, or consume such models to perform downstream tasks, to be aware of the presence of annotation artifacts, and adjust their assessments of model performance accordingly. It is our hope that these findings can be used to inform error analysis and improve predictive models that inform patient care. 

\section*{Acknowledgments}
We thank the four anonymous reviewers whose feedback and suggestions helped improve this manuscript. The first author was supported by the National Institute of Standards and Technology's (NIST) Professional Research Experience Program (PREP). This research was also supported by the DARPA KAIROS program. The views and conclusions contained in this publication are those of the authors and should not be interpreted as representing official policies or endorsements of NIST, DARPA, or the U.S. Government.

\bibliographystyle{acl2021}
\bibliography{anthology,acl2021}

\pagebreak

\appendix
\section{Appendix}
\label{sec:appendix}
\subsection{Hypothesis-only Baseline Analysis}
To conduct the analysis presented in \hyperref[sec:mednliContainsArtifacts]{Section 3}, we take the MedNLI training dataset as input, and exclude the premise text for each training example. We cast the text of each training hypothesis to lowercase, but do not perform any additional preprocessing. We use an off-the-shelf \texttt{fastText} classifier, with all model hyperparameters set to their default values with the exception of \texttt{wordNgrams}, which we set equal to 2 to allow the model to use bigrams in addition to unigrams~\cite{joulin2016bag}. We evaluate the trained classifier on the hypotheses contained in the MedNLI dev and test datasets, and report results for each split.

\subsection{Lexical Artifact Analysis}
To perform the analysis presented in \hyperref[sec:lexicalArtifacts]{Section 4}, we cast each hypothesis string in the MedNLI training dataset to lowercase. We then use a \texttt{scispaCy} model pre-trained on the \texttt{en\_core\_sci\_lg} corpus for tokenization and clinical named entity recognition (CNER)~\citep{neumann2019scispacy}. Next, we merge multi-token entities, using underscores as delimiters---e.g., ``brain injury'' $\rightarrow$ ``brain\_injury''. 

When computing token-class pointwise mutual information (PMI), we exclude tokens that appear less than five times in the overall training dataset's hypotheses. Then, following \citet{gururangan2018annotation}, who apply add-100 smoothing to raw counts to highlight particularly discriminative token-class co-occurrence patterns, we apply add-50 smoothing to raw counts. Our approach is similarly motivated; our choice of 50 reflects the smaller state space associated with a focus on the clinical domain. 
\subsection{Semantic Analysis of Heuristics}
To perform the statistical analysis presented in \hyperref[sec:annotatorHeuristics]{Section 5}, we take the premise-hypothesis pairs from the MedNLI training, dev, and test splits, and combine them to produce a single corpus. We use a \texttt{scispaCy} model pre-trained on the \texttt{en\_core\_sci\_lg} corpus for tokenization and entity linking~\cite{neumann-etal-2019-scispacy}, and link against the Medical Subject Headings (MeSH) knowledge base. We take the top-ranked knowledge base entry for each linked entity. Linking against MeSH provides a unique concept id, canonical name, alias(es), a definition, and one or more MeSH tree numbers for each recovered entity. Tree numbers convey semantic type information by embedding each concept into the broader MeSH hierarchy~\footnote{\url{https://meshb.nlm.nih.gov/treeView}}. We operationalize each of our heuristics with a set of MeSH-informed semantic properties, which are defined as follows:

\begin{enumerate}
    \item \textbf{Hypernym Heuristic:} a premise-hypothesis pair satisfies this heuristic if specific clinical concept(s) appearing in the premise appear in a more general form in the hypothesis. Formally:
    $\{ (p,h) | \upvarphi(p) \subsetneq \upvarphi(h)\}$.
    MeSH tree numbers are organized hierarchically, and increase in length with specificity. Thus, when a premise entity and hypothesis entity are left-aligned, the hypothesis entity is a hypernym for the premise entity if the hypothesis entity is a substring of the premise entity. To provide a concrete example: \emph{diabetes mellitus} is an \emph{endocrine system disease}; the associated MeSH tree numbers are C19.246 and C19, respectively.
    \item \textbf{Probable Cause Heuristic:} a premise-hypothesis pair satisfies this heuristic if: (1) the premise contains one or more MeSH entities belonging to high-level categories C (diseases), D (chemicals and drugs), E (analytical, diagnostic and therapeutic techniques, and equipment) or F (psychiatry and psychology); and (2) the hypothesis contains one or more MeSH entities that can be interpreted as providing a plausible causal or behavioral explanation for the condition, finding, or event described in the premise (e.g., smoking, substance-related disorders, mental disorders, alcoholism, homelessness, obesity). 
    \item \textbf{Everything Is Fine Heuristic:} a premise-hypothesis pair satisfies this heuristic if the hypothesis contains one or more of the same MeSH entities as the premise (excluding the \emph{patient} entity, which appears in almost all notes) and also contains: (1) a negation word or phrase (e.g., \emph{does not have}, \emph{no finding}, \emph{no}, \emph{denies}); or (2) a word or phrase that affirms the patient's health (e.g., \emph{normal}, \emph{healthy}, \emph{discharged}).
\end{enumerate}

For each heuristic, we subset the complete dataset to find pattern-satisfying premise-heuristic pairs. We use this subset when performing the $\chi^2$ tests. 

\subsection{Adversarial Filtering}
When implementing \texttt{AFLite}, we follow ~\citet{sakaguchi2020winogrande}. We use a smaller training set size of $m = 5620$, but keep the remaining hyperparameters unchanged, such that the ensemble consists of $n = 64$ logistic regression models, the filtering cutoff, $k = 500$, and the filtering threshold $\tau= 0.75$.
\end{document}